\documentclass[10pt,twocolumn,letterpaper]{article}

\usepackage[pagenumbers]{cvpr} % To force page numbers, e.g. for an arXiv version

\usepackage[dvipsnames]{xcolor}

\definecolor{cvprblue}{rgb}{0.21,0.49,0.74}
\usepackage[pagebackref,breaklinks,colorlinks,citecolor=cvprblue]{hyperref}
\usepackage[dvipsnames]{xcolor}
\usepackage[outercaption]{sidecap}    
\usepackage{color, colortbl}
\usepackage{multirow}
\usepackage{makecell}
\usepackage{pifont}% http://ctan.org/pkg/pifont
\usepackage{arydshln}
\usepackage{marvosym}
\newcommand{\cmark}{\ding{51}}%
\newcommand{\xmark}{\ding{55}}%
\definecolor{golden}{RGB}{255, 215, 0} % RGB values for golden

\newcommand{\AlgoName}{QA-ViT }
\newcommand{\AlgoNameNoSpace}{QA-ViT}

\newcommand*\samethanks[1][\value{footnote}]{\footnotemark[#1]}
\def\paperID{8234} % *** Enter the Paper ID here
\def\confName{CVPR}
\def\confYear{2024}

\title{Question Aware Vision Transformer for Multimodal Reasoning}

\author{
Roy Ganz\thanks{Work done during an Amazon internship.}\\
Technion, Israel\\
{\tt\small ganz@cs.technion.ac.il}
\and
Yair Kittenplon\thanks{Corresponding author.}\\
AWS AI Labs\\
{\tt\small yairk@amazon.com}
\and
Aviad Aberdam\\
AWS AI Labs\\
{\tt\small aaberdam@amazon.com}
\and
Elad Ben Avraham\\
AWS AI Labs\\
{\tt\small eladba@amazon.com}
\and
Oren Nuriel\\
AWS AI Labs\\
{\tt\small onuriel@amazon.com}
\and
Shai Mazor\\
AWS AI Labs\\
{\tt\small smazor@amazon.com}
\and
Ron Litman\samethanks\\
AWS AI Labs\\
{\tt\small litmanr@amazon.com}
}

\begin{document}
\maketitle
\begin{abstract}
Vision-Language (VL) models have gained significant research focus, enabling remarkable advances in multimodal reasoning. 
These architectures typically comprise a vision encoder, a Large Language Model (LLM), and a projection module that aligns visual features with the LLM's representation space. 
Despite their success, a critical limitation persists: the vision encoding process remains decoupled from user queries, often in the form of image-related questions. 
Consequently, the resulting visual features may not be optimally attuned to the query-specific elements of the image.
To address this, we introduce \AlgoNameNoSpace, a Question Aware Vision Transformer approach for multimodal reasoning, which embeds question awareness directly within the vision encoder.
This integration results in dynamic visual features focusing on relevant image aspects to the posed question.
\AlgoName is model-agnostic and can be incorporated efficiently into any VL architecture.
Extensive experiments demonstrate the effectiveness of applying our method to various multimodal architectures, leading to consistent improvement across diverse tasks and showcasing its potential for enhancing visual and scene-text understanding.
\end{abstract}    
\vspace{-4mm}
\section{Introduction}
\label{sec:intro}

\begin{figure}[t]
    \centering
    \includegraphics[width=0.98\linewidth]{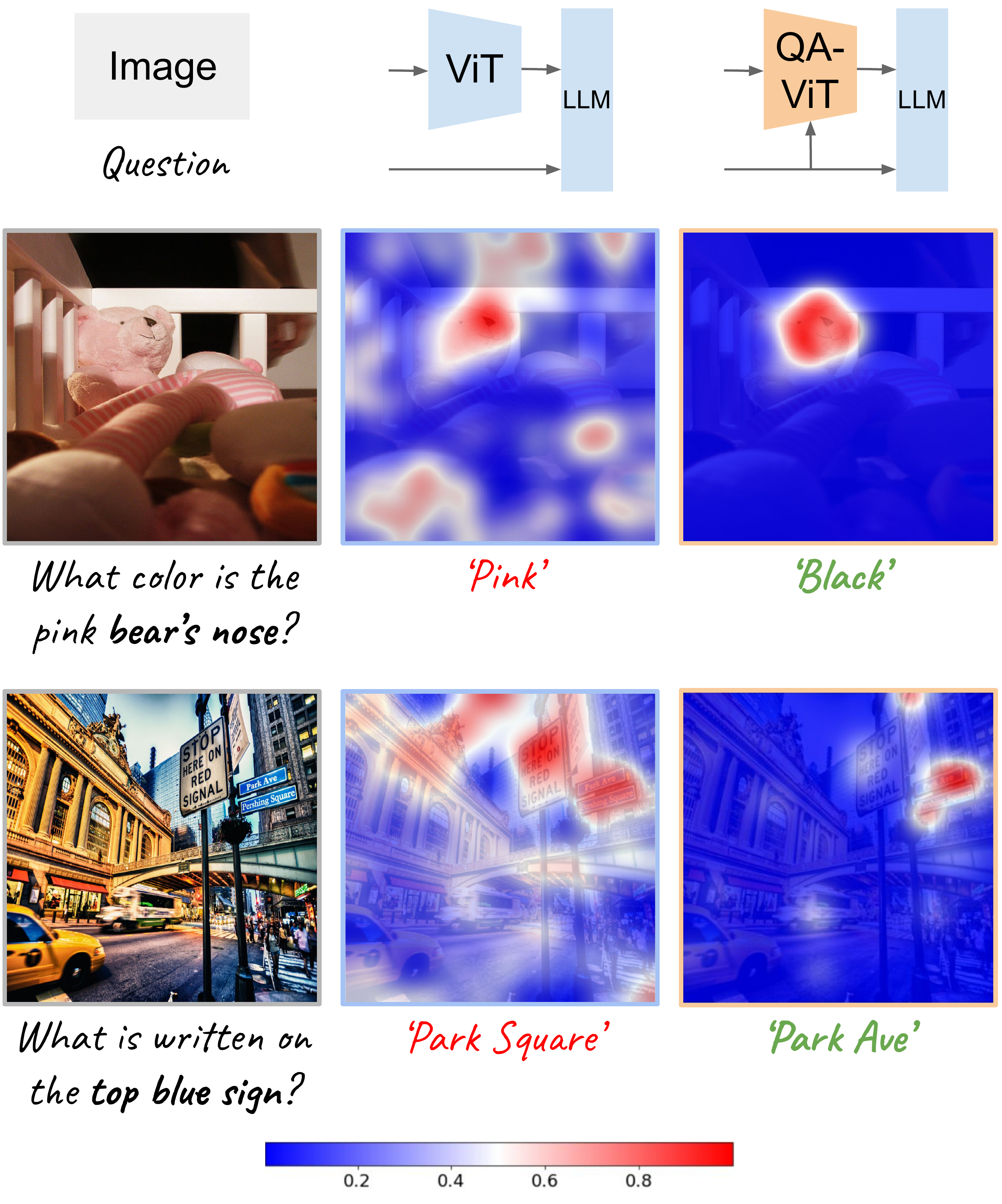}
    \caption{\textbf{Question-Aware Vision Encoding.} Comparative illustrations for VQAv2 (upper) and TextVQA (lower) predictions of ViT+T5 and QA-ViT+T5 VL models. Employing GradCAM highlights the focus areas with respect to key terms in the posed questions. This vividly demonstrates the motivation behind \AlgoNameNoSpace: enhancing ViT with the question enables it to focus on the relevant image aspects, resulting in more accurate predictions.}
    \vspace{-4mm}
    \label{fig:teaser}
\end{figure}

% high level
In recent years, VL architectures have emerged as a pivotal research area, leading to significant progress in the domain of multimodal reasoning~\cite{li2022blip,alayrac2022flamingo,li2023blip,liu2023visual,dai2023instructblip,zhang2023llavar,hu2023bliva,ganz2023towards,rotstein2023fusecap,ganz2023clipag}. 
Such architectures fundamentally seek to bridge the gap between visual and textual data, enabling models to interpret, comprehend, and generate content based on both visual and textual information. 
This fusion of modalities has diverse applications and tasks, from image captioning (CAP)~\cite{chen2015microsoft,sidorov2020textcaps} and visual question answering (VQA)~\cite{antol2015vqa,singh2019towards} to tasks in autonomous robotics and human-computer interactions.
As the list of applications continues to grow, the role of VL architectures becomes increasingly crucial within the broader field of deep learning.

% Vision-Language Modeling
At the heart of multimodal VL architectures lies the concept of vision-language Modeling. These models typically consist of three essential steps. First, a unimodal vision architecture extracts meaningful information from images. Typically, the vision encoder is a frozen Vision-Transformer (ViT), often based on CLIP~\cite{dosovitskiy2020image,radford2021learning}. Second, a projection module bridges the gap between vision and language, transforming visual features into ones that can be comprehended and processed by a language model. This module is usually either a simple linear layer or MLP~\cite{liu2023visual,liu2023improved,zhang2023llavar}, or a cross-attention-based transformer architecture~\cite{li2023blip,dai2023instructblip,bai2023qwen}. Lastly, the projected visual information and the textual instruction, commonly in the form of questions or prompts, are inserted into a Large Language Model (LLM) to complete the task.

% Problem and Limitations
Despite the remarkable progress achieved in VL research, we have identified an intriguing yet often overlooked limitation within such architectures.
The success of such a model hinges on its ability to not only comprehend the visual content but also to do so through the lens of the accompanying textual instruction, \textit{e.g.}, the provided question, often requiring focus on fine-grained details inside the entire image.
Existing architectures, however, are suboptimal in this aspect, as they perform the vision encoding unaware of the posed question, resulting in visual features not optimally aligned with the user query.
As the vision encoder outputs a fixed size features sequence $F_{V}$, it is limited in the level of information encoded in them.
Due to the relatively high abstraction level, it is likely to disregard or overlook low-level details in the image. This oversight becomes particularly problematic in scenarios where nuanced image understanding is essential to accurately respond to queries.
Thus, we claim that the vision encoder $\mathcal{V}$ should be cast from a single input function into a conditional function.
Namely, $\mathcal{V}(I | Q)$ instead of $\mathcal{V}(I)$, where $I, Q$ are the image and question, respectively.

\begin{figure}[t]
    \centering
    \includegraphics[width=0.98\linewidth]{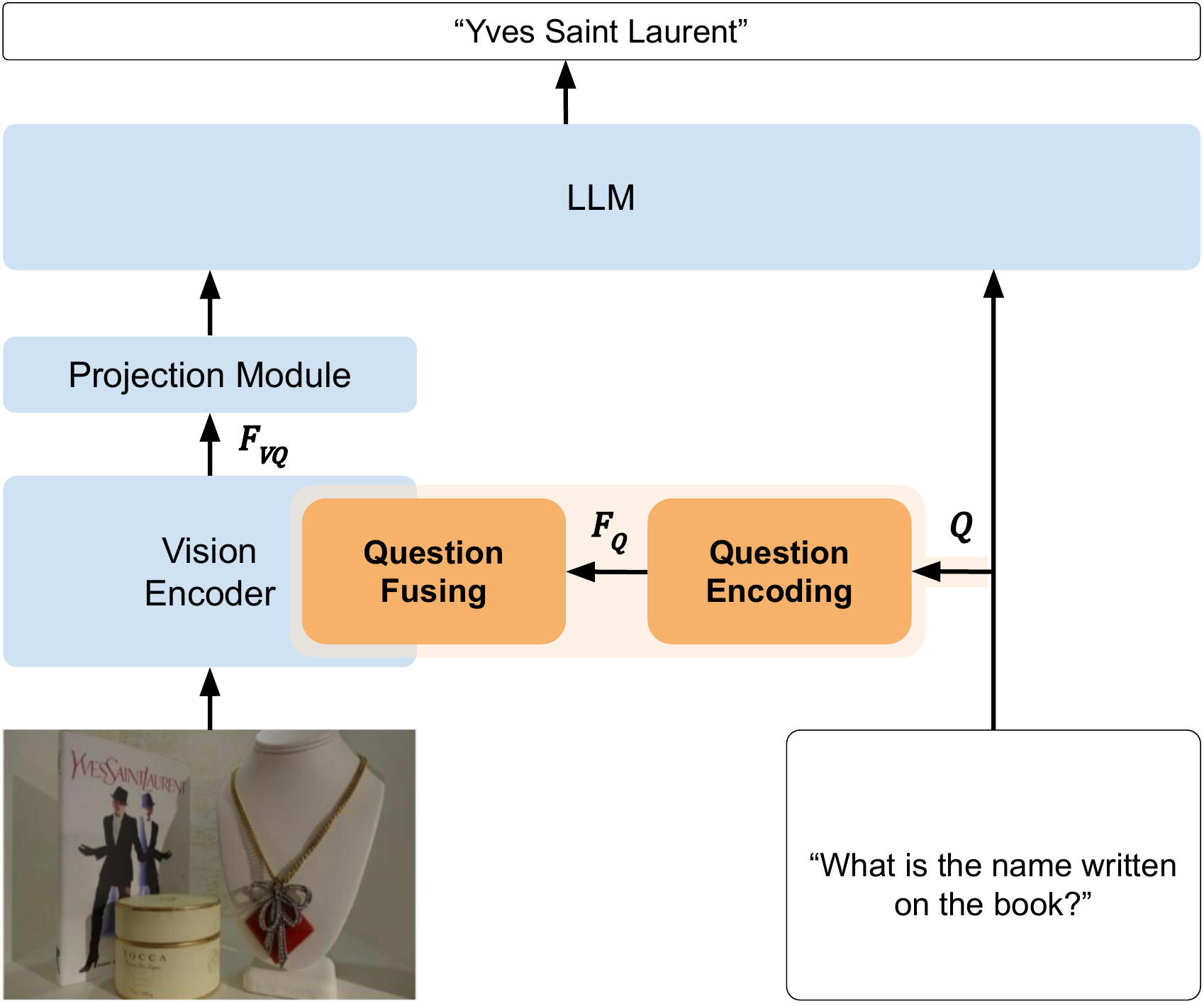}
    \caption{\textbf{Method overview.} 
    A high-level illustration of the \AlgoName (highlighted in orange) incorporated into a general VL architecture (depicted in blue).
    This is achieved by encoding the question $Q$ into features $F_Q$, which are fused into the vision encoder, resulting in question-aware visual features $F_{VQ}$.
    }
    \vspace{-4mm}
    \label{fig:overview}
\end{figure}

% Proposed Solution
To mitigate this limitation and yield a textual conditioned vision encoding, we present \textbf{\AlgoNameNoSpace}, Question Aware Vision Transformer for multimodal reasoning. 
The intuition of our method is clear: if the model understands the posed question and the inherent context, it can extract visual features that directly correspond to the relevant image aspects essential for answering it correctly.
We illustrate this behavior in \cref{fig:teaser}; By applying GradCAM~\cite{selvaraju2017grad} to both vanilla CLIP-based ViT and \AlgoNameNoSpace, w.r.t. textual prompts correspond with a distinct spatial location.
While the baseline tends to favor high abstraction level features, even when prompted with region-specific descriptions, \AlgoName focuses significantly more on the relevant image parts.
For instance, considering the bottom image and the question like ``What is written on the top blue sign?'', we can see that while the baseline vision encoder generates features that contain a wealth of information about the scene (\textit{e.g.}, the buildings, cars, and people), \AlgoName is able to pinpoint the specific region of interest, namely, the blue sign.
Our approach achieves the above goal by directly integrating textual representations into any vision encoder while keeping most of it frozen, preserving its visual understanding capabilities (\cref{fig:overview}).
In practice, we utilize the preexisting self-attention mechanism in the ViT to also attend to textual encodings, representing the user query.

% Experiments
To demonstrate \AlgoName effectiveness, we leverage the model-agnostic nature of our method and integrate it into top-performing systems, including BLIP2~\cite{li2023blip}, InstructBLIP~\cite{dai2023instructblip}, and LLaVA-1.5~\cite{liu2023improved}. 
In addition, we also integrate \AlgoName into a simple ViT+T5 architecture, without pretraining, to demonstrate its benefit when training an unaligned VL system from scratch.
We train all these architectures on a combined dataset of visual question answering and image captioning, requiring visual and Optical Character Recognition (OCR) understanding, and evaluate them accordingly.
Despite the architectural differences between the considered VL models in the vision-encoder, projection module (QFormer vs. MLP), and LLM structure (encoder-decoder vs. decoder only), extensive experiments show that \AlgoName consistently improves the performance over all the tested models and benchmarks, attesting to its versatility.

To summarize:
\begin{itemize}[nolistsep,leftmargin=*]
    \item We identify an overlooked suboptimality in the paradigm of vision-language modeling stemming from the lack of instruction-aware image encoding.
    \item We introduce \AlgoNameNoSpace, a model-agnostic method that enables existing vision encoders to be conditioned on textual prompts or questions.
    \item Thorough experiments on multiple architectures demonstrate our method's ability to enhance multimodal reasoning, improving the performance on various benchmarks.
\end{itemize}

\section{Related Work}
\label{sec:bg}
\paragraph{Vision-Language Models.}

Earlier-generation VL models pursue the paradigm of rigorous and extensive pretraining, using contrastive losses, followed by designated fine-tuning for specific tasks~\cite{li2022blip,wang2022git,albef,wang2022ofa,li2022mplug,yang2021tap}. 
While this approach constituted a critical milestone, it led to specialist models that only perform well on a specific downstream task~\cite{singh2019towards,biten2019scene,ganz2023towards}.
By leveraging the capabilities of recent Large Language Models (LLMs)~\cite{flan-t5,taori2023stanford,touvron2023llama,touvron2023llama2},
current top-performing VL models are generalist models, showcasing remarkable performance across various VL tasks. Interestingly, such models demonstrate strong zero-shot performance and generalization to unseen data and tasks~\cite{alayrac2022flamingo,li2023blip,dai2023instructblip,liu2023improved,bai2023qwen,chen2023pali}, and sometimes even surpassing specialist models. 

Architecturally, there are two main types of VL models, which mainly differ in the integration mechanism of the visual features into the LLM.
The first type projects the visual features using a cross-attention-based transformer model (\textit{e.g.}, QFormer), which also reduces the visual sequence length~\cite{li2023blip,dai2023instructblip,bai2023qwen}. 
The introduction of such a mechanism enables keeping both the LLM and the vision encoder frozen. 
The second line of research demonstrates that the projection module can be simplified to a linear projection (or an MLP) while also training the LLM~\cite{liu2023visual,zhang2023llavar,liu2023improved,chen2023pali}.
Despite such differences, all current top-performing VL models perform image encoding in an unaware manner to the given textual prompt. 

\paragraph{Question-Aware Vision Encoding.}

A possible solution for the limitation above was proposed in the OCR-free text-oriented multimodal understanding by pix2struct~\cite{lee2023pix2struct}, which suggests directly rendering the question as a header at the top of the original image instead of passing it to the LLM.
However, this approach relies highly on their OCR-oriented pretraining and is suboptimal in the general VL case.
Another step towards instruction-aware visual features is InstructBlip~\cite{dai2023instructblip}, which introduces the visual features into the QFormer alongside the instruction.
Nevertheless, it operates solely on top of the outputs of the vision encoder and, thus, is incapable of compensating for overlooked image aspects.  
In this paper, we propose to integrate question information into any ViT-based image encoder in a flexible and modular manner.

\section{Method}
\label{sec:method}

Our method proposes a versatile and lightweight model-agnostic approach, which can be integrated into any vision transformer model in any VL architecture, designed to transform trained image encoders into question-aware ones effectively.
Formally, given the image and question $I, Q$, we argue that the vision encoding module $\mathcal{V}$ should be casted into a conditioned one:
\begin{equation}
    F_V = \mathcal{V}(I) \rightarrow F_{VQ} = \mathcal{V}(I | Q).
\end{equation}
In this section, we first describe our high-level design and then delve into the details of each building block.

\subsection{Overall Architecture}
As illustrated in \cref{fig:overview}, our method comprises two fundamental components.
First, the question, denoted as $Q$, is fed into a ``\textbf{Question Encoding}'' module, which processes and projects the textual prompt, bridging the gap between the linguistic and visual features domains.
Subsequently, the textual encoded features, denoted as $F_Q$, are integrated inside a frozen vision model via ``\textbf{Question Fusing}'' module, producing text-aware visual features $F_{VQ}$.
Lastly, the $F_{VQ}$ is projected by the projection module, concatenated with the instruction embeddings, and fed into the LLM, which processes and produces the overall system's output.
In general, \AlgoName modifies solely the vision encoder, maintaining the rest of the architecture intact.

\subsection{Question Encoding}
In order to introduce text prompts $Q$ into an unimodal vision transformer, we propose a streamlined two-stage process.
\paragraph{Question Representation.}
\vspace{-2mm}
First, we encode the natural language prompt (\textit{e.g.}, the question) into meaningful representations, denoted as $F_Q'$.
Formally, we define this operation as $\mathcal{E}(Q) = F_Q'$, where 
$\mathcal{E}$ represents the encoding function.
This step introduces flexibility in choosing $\mathcal{E}$, the source of these textual representations -- the preexisting LLM's encoder or embeddings or a designated language model.
We mainly focus on the former as it offers more parameter efficiency and can lead to more seamless integration, as the same LLM subsequently processes the visual features. 
We compare these approaches in \cref{sec:design_ablation}.

\paragraph{Representation Projection.}
\vspace{-2mm}
Second, we utilize MLPs to project the textual representations into the vision model features space.
Due to the vision model's hierarchical structure, different layers have different abstraction levels~\cite{dosovitskiy2020image,raghu2021vision}. Hence, we adopt a per-layer MLP to obtain better alignment. 
We denote the projected textual representation for layer $i$ as $F^{i}_Q$.
Overall, the question encoding phase operates as follows:
\begin{equation}
    F^i_Q = \operatorname{MLP}^i(\mathcal{E}(Q)).
\end{equation}
For simplicity, we omit the layer index from now on.

\subsection{Question Fusing}
\begin{figure}[t]
    \centering
    \includegraphics[width=\linewidth]{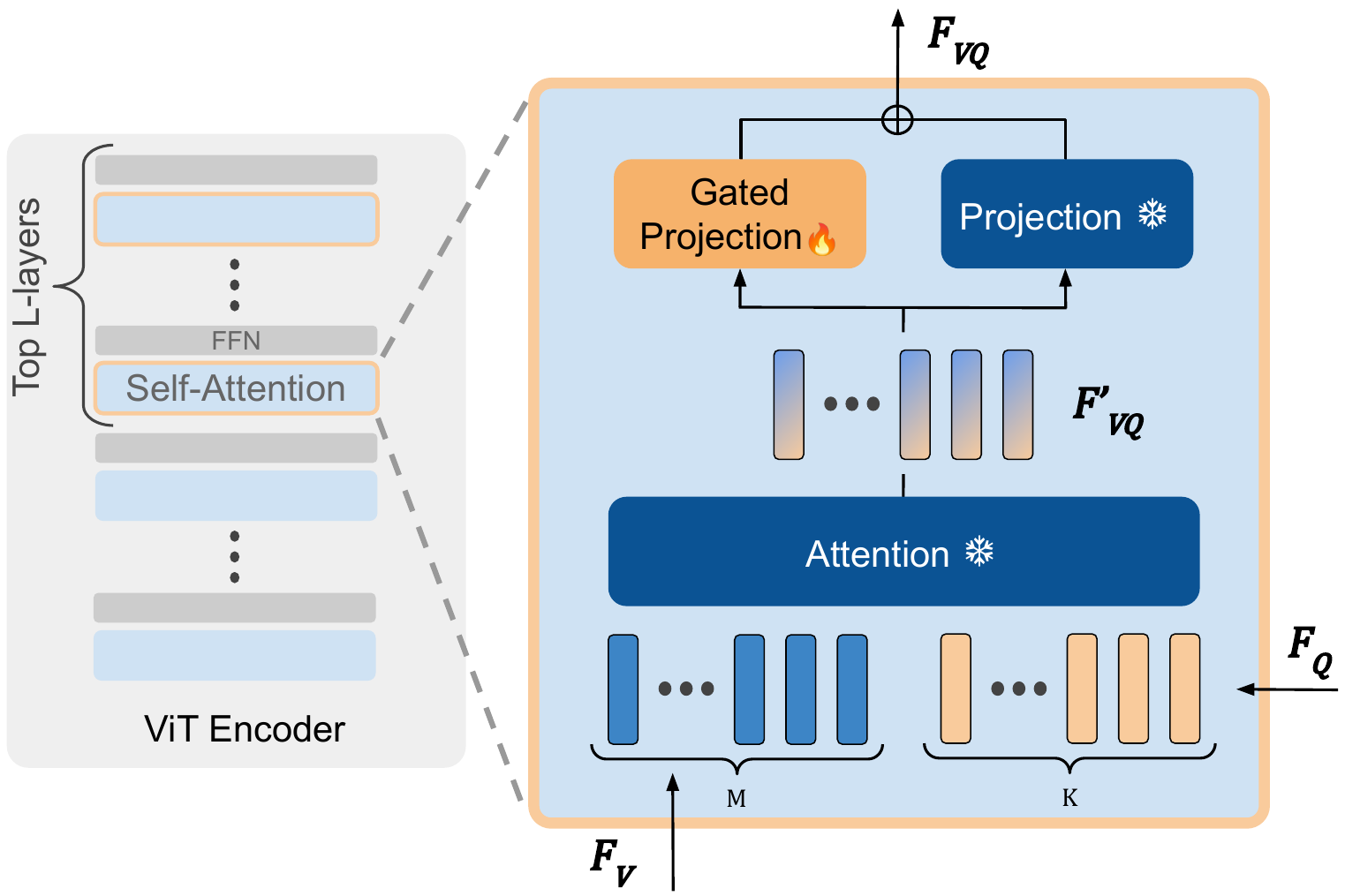}
    \caption{\textbf{Textual representations fusing.}
    Left: General scheme of the ViT encoder. Right: Zoom in to our fusing mechanism in one of the top-L self-attention layers. The $M$ visual features from the previous layer $F_{V}$,  are concatenated with $K$ textual features $F_Q$ and fed into the frozen self-attention mechanism to obtain $M$ text-attended visual representations $F_{VQ}'$. Next, a parallel gated projection obtains the question-aware visual features of $F_{VQ}$.
    }
    \label{fig:fusing}
\end{figure}

Given the projected textual representations $F_Q$, we propose a parameter-efficient fusing mechanism to integrate them into frozen ViT architectures in a model-agnostic way.
Keeping the vision encoder frozen enables text-conditioned encoding of the image while preserving the model's original capabilities intact.
While such integration can be done in various ways, we propose a straightforward approach that harnesses the ViT preexisting self-attention mechanism, illustrated in \cref{fig:fusing}.

\paragraph{Fusing Mechanism.}
\vspace{-2mm}
We extend the input sequence of the self-attention layer to contain the projected representations ${F_{Q}\in \mathbb{R}^{K \times C}}$ by concatenating it with the visual representations ${F_{V}\in \mathbb{R}^{M \times C}}$, where $C$ is the channel dimension.
This yields a sequence of length ${K + M}$, containing vision and question information. 
Next, the frozen self-attention mechanism is applied to produce the attention scores and outputs while also attending to the textual information $F_{Q}$, enabling cross-modal attention.
We select the attention output that corresponds with the input visual representations, resulting in ${F'_{VQ}\in \mathbb{R}^{M\times C}}$.
More formally,
\begin{equation}
    F'_{VQ} = \operatorname{Attention}(\operatorname{concat}(F_V,F_Q))_{[\operatorname{0:M}]}.
\end{equation}

An additional projection followed by a learnable gating mechanism~\cite{hochreiter1997long,alayrac2022flamingo,ganz2023towards,aberdam2023clipter} is introduced in parallel to the existing frozen projection head.
This module compensates for the distribution shift from incorporating question information in the frozen self-attention layer.
% to compensate for the distribution shift caused by including the question information in the frozen self-attention layer.
The goal of such a gating is to enable the gradual blending of the residual projected information with the existing one, avoiding a significant feature modification and a degradation of the overall performance. Such gating is done by multiplying the additional projection layer's outputs with $\tanh(\beta)$, where $\beta$ is a learnable parameter initialized to zero.
This technique is designed to maintain the layer's outputs with minimal deviation at initialization, improving stability while enabling a residual learnable stream of information.
Mathematically, our fusing mechanism functions as follows:
\begin{equation}
    F_{VQ} = \mathcal{P}(F'_{VQ}) + \mathcal{P}_{g}(F'_{VQ}) \cdot \tanh(\beta).
\end{equation}

% Specifically, given the attended features $F'_{VQ}\in \mathbb{R}^{M\times C}$ and projection matrices $P, P_g \in {R}^{C\times C}$, the output of the self-attention layer is set to be $F_{VQ}=F'_{VQ}\cdot(P + P_g \cdot \tanh(\beta))$, yielding a residual learnable stream of information.

\paragraph{Integration Point.}
\vspace{-2mm}
An important design choice in our fusing mechanism is the choice of the integration point of the textual representations into the vision transformer layers. 
Specifically, we perform \texttt{late fusion}, namely, applying the fusing in the top $L$ self-attention layers of the $N$-layered ViT, where ${L<N}$.
This choice is motivated by the nature of ViT layers hierarchy -- lower layers primarily capture low-level visual details, while the higher layers mainly focus on high-level concepts~\cite{dosovitskiy2020image,raghu2021vision}.
% Specifically, these higher layers have a more abstract understanding of the image, focusing on contextual relationships and encapsulating entire object concepts.
Therefore, the likelihood of disregarding fine-grained details is expected to emerge in the higher layers, making them an optimal target for our method. We validate this choice in \cref{sec:ablation}.

\begin{figure*}[t]
    \centering
    \includegraphics[width=0.98\textwidth]{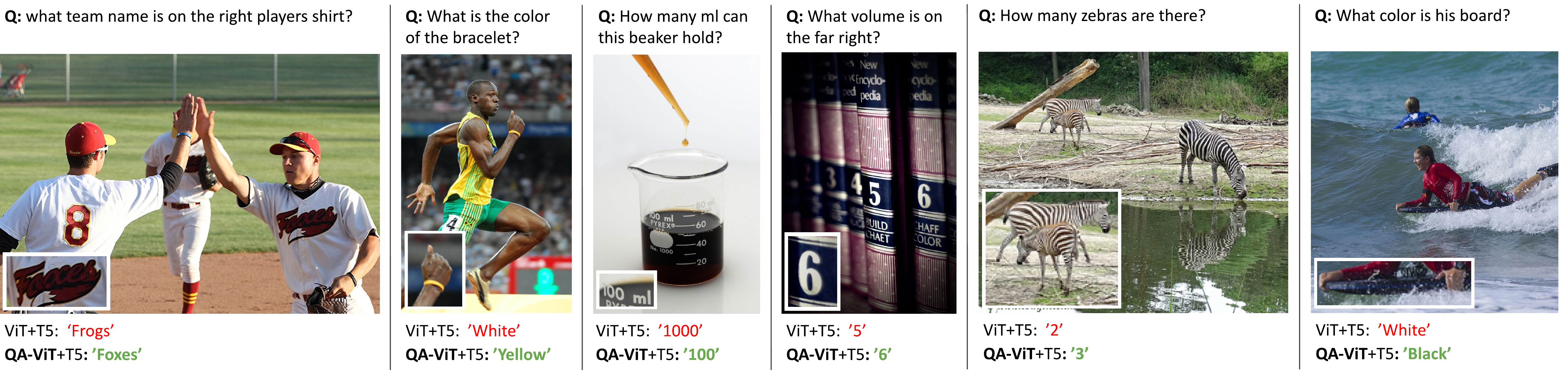}
    \caption{\textbf{Paying attention to details in visual question answering.}
    Representative examples require answering questions regarding subtle or less conspicuous image details (zoomed-in) from VQAv2 and TextVQA datasets. 
    Each sample includes an image-question pair alongside predictions from ViT+T5 and \AlgoNameNoSpace +T5, where green indicates correct predictions and red indicates incorrect ones.
    }
    \vspace{-3mm}
    \label{fig:exmaples}
\end{figure*}

\section{Experiments}
\label{sec:exps}

We conduct a comprehensive set of experiments to assess the capabilities of \AlgoNameNoSpace. 
Given the model-agnostic nature of our method, which enables seamless integration into any existing VL architecture, our experiments are designed to showcase its versatility in two distinct architectural settings.
In the first setting, we experiment with a straightforward VL approach consisting of a vision encoder and encoder-decoder-based LLM, denoted as ViT+T5.
The second setting involves integrating our method into already trained top-performing vision-language models, specifically LLAVA-1.5~\cite{liu2023improved}, BLIP2~\cite{li2023blip}, and instructBLIP~\cite{dai2023instructblip}. This allows us to assess the benefits of \AlgoName for already finetuned models.
In both settings, we train and evaluate the models using a combined dataset of visual question answering and image captioning, requiring both visual and OCR understanding~\cite{aberdam2022multimodal,litman2020scatter,aberdam2023clipter}.
In the OCR case, we are interested in the \textbf{OCR-free} setting; we do not equip the models with OCR tokens.

\subsection{Training Data}
\label{sec:data}
For training across all considered architectures, we adopt a multi-task approach using concatenated VL datasets that involve reasoning over both visual and OCR information.
In particular, we consider general visual question-answering datasets~\cite{balanced_vqa_v2,krishna2017visual} alongside scene-text~\cite{singh2019towards,biten2019scene,mishra2019ocr} and document-oriented ones~\cite{mathew2021docvqa,mathew2022infographicvqa,masry2022chartqa}. 
For these datasets, We insert the question representations into the vision encoder when applying \AlgoNameNoSpace.
In addition, we include captioning datasets (COCO Captions~\cite{coco_cap} and TextCaps~\cite{sidorov2020textcaps}), which leads to additional improvements, as can be seen in \cref{sec:ablation_data}).
In the captioning data, we utilize a random template instruction, as in \cite{dai2023instructblip}, \textit{e.g., ``Please provide a short depiction of the picture''} and insert them into the ViT.
We provide the complete list of such templates in the supplementary materials, alongside further details on the training dataset composition.
Overall, our dataset comprises approximately $3$ million assets from multiple training datasets of different sizes. We adopt a sampling strategy proportional to each dataset's size during training to address the size disparity. This approach is designed to prevent overfitting smaller datasets and underfitting larger ones.

\begin{table*}[t]
  \centering
  \bgroup
  \def\arraystretch{1.1}
  \resizebox{1\linewidth}{!}{%
  \begin{tabular}{ll|cccccc|cc}
    \toprule
    \multirow{3}{*}{\textbf{Method}} & \multirow{3}{*}{LLM} & \multicolumn{2}{c}{\textbf{General}} & \multicolumn{3}{c}{\textbf{Scene-Text}} & \textbf{0-shot} & \multicolumn{2}{c}{\textbf{Average}} \\
    & & $\text{VQA}^{\text{v2}}$ & COCO & $\text{VQA}^{\text{T}}$ & $\text{VQA}^{\text{ST}}$ & TextCaps & VizWiz & \multirow{2}{*}{General} & \multirow{2}{*}{Scene-Text}\\
    & & \textcolor{gray}{\small vqa-score} & \textcolor{gray}{\small CIDEr} & \textcolor{gray}{\small vqa-score} & \textcolor{gray}{\small ANLS} & \textcolor{gray}{\small CIDEr} & \textcolor{gray}{\small vqa-score} & & \\
    % \hline
    % \rowcolor{gray!30}  % Set the background color of the second row
    % ViT+T5-small   & Flan-T5-small     & 58.7 & 101.7 & 32.8 & 40.4 & 74.0 & 19.6 & 80.2 & 55.3\\
    % + \AlgoName   &                    & 65.3  & 109.0 & 39.1 & 47.6 & 86.9 & 22.2 & 87.2 & 65.1\\
    % $\Delta$     &  & {\color{OliveGreen}\textbf{+6.6}} & {\color{OliveGreen}\textbf{+7.3}} & {\color{OliveGreen}\textbf{+6.3}} & {\color{OliveGreen}\textbf{+7.2}} & {\color{OliveGreen}\textbf{+12.9}} & {\color{OliveGreen}\textbf{+2.6}} & {\color{OliveGreen}\textbf{+7.0}} & {\color{OliveGreen}\textbf{+9.8}}\\
    \hline
    \rowcolor{gray!30}  % Set the background color of the second row
    ViT+T5-base   & Flan-T5-base     & 66.5 & 110.0 & 40.2 & 47.6 & 86.3 & 23.7 & 88.3 & 65.1 \\
    + \AlgoName   &                  & 71.7 & 114.9 & 45.0 & 51.1 & 96.1 & 23.9 & 93.3 & 72.1 \\
    $\Delta$     &  & {\color{OliveGreen}\textbf{+5.2}} & {\color{OliveGreen}\textbf{+4.9}} & {\color{OliveGreen}\textbf{+4.8}} & {\color{OliveGreen}\textbf{+3.5}} & {\color{OliveGreen}\textbf{+9.8}} & {\color{OliveGreen}\textbf{+0.2}} & {\color{OliveGreen}\textbf{+5.0}}& {\color{OliveGreen}\textbf{+7.0}}\\
    \hline
    \rowcolor{gray!30}  % Set the background color of the second row
    ViT+T5-large   & Flan-T5-large     & 70.0 & 114.3 & 44.7 & 50.6 & 96.0 & 24.6 & 92.2 & 71.8 \\
    + \AlgoName    &                   & 72.0 & 118.7 & 48.7 & 54.4 & 106.2 & 26.0 & 95.4 & 78.9 \\
    $\Delta$     &  & {\color{OliveGreen}\textbf{+2.0}} & {\color{OliveGreen}\textbf{+4.4}} & {\color{OliveGreen}\textbf{+4.0}} & {\color{OliveGreen}\textbf{+3.8}} & {\color{OliveGreen}\textbf{+10.2}} & {\color{OliveGreen}\textbf{+1.4}} & {\color{OliveGreen}\textbf{+3.2}} & {\color{OliveGreen}\textbf{+7.1}}\\
    \hline
    \rowcolor{gray!30}  % Set the background color of the second row
    ViT+T5-xl     & Flan-T5-xl     & 72.7 & 115.5 & 48.0 & 52.7 & 103.5 & 27.0 & 94.1 & 77.0 \\
    + \AlgoName   &                & 73.5 & 116.5 & 50.3 & 54.9 & 108.2 & 28.3 & 95.0 & 80.4\\
    $\Delta$      &  & {\color{OliveGreen}\textbf{+0.8}} & {\color{OliveGreen}\textbf{+1.0}} & {\color{OliveGreen}\textbf{+2.3}} & {\color{OliveGreen}\textbf{+2.2}} & {\color{OliveGreen}\textbf{+4.7}} & {\color{OliveGreen}\textbf{+1.3}} & {\color{OliveGreen}\textbf{+0.9}} & {\color{OliveGreen}\textbf{+3.4}}\\
    % \hline
    % \hline
    \Xhline{1pt}
    \rowcolor{gray!30}  % Set the background color of the second row
    BLIP2~\cite{li2023blip}   & Flan-T5-xl     & 72.5 & 134.8 & 34.5 & 36.4 & 93.6 & 28.2 &  103.7 & 64.5 \\
    + \AlgoName &            & 74.6 & 136.6 & 36.6 & 38.1 & 97.4 & 28.4 & 105.6 & 67.4 \\
    $\Delta$     &  & {\color{OliveGreen}\textbf{+2.1}} & {\color{OliveGreen}\textbf{+1.8}} & {\color{OliveGreen}\textbf{+2.1}} & {\color{OliveGreen}\textbf{+1.7}} & {\color{OliveGreen}\textbf{+3.8}} & {\color{OliveGreen}\textbf{+0.2}} & {\color{OliveGreen}\textbf{+1.9}} & {\color{OliveGreen}\textbf{+2.9}}\\
    \hline
    \rowcolor{gray!30}  % Set the background color of the second row
    BLIP2~\cite{li2023blip}       & Flan-T5-xxl & 74.8 & 134.8                  & 36.5 & 37.9 & 97.4 & 29.8 & 104.8 & 67.3 \\
    + \AlgoName &             & 75.6 & 135.9 & 37.5 & 39.9 & 98.7 & 30.4 & 105.8 & 68.7 \\
    $\Delta$     &  & {\color{OliveGreen}\textbf{+0.8}} & {\color{OliveGreen}\textbf{+1.1}} & {\color{OliveGreen}\textbf{+1.0}} & {\color{OliveGreen}\textbf{+2.0}} & {\color{OliveGreen}\textbf{+1.3}} & {\color{OliveGreen}\textbf{+0.6}} &
    {\color{OliveGreen}\textbf{+1.0}} & {\color{OliveGreen}\textbf{+1.4}}\\
    \hline
    \rowcolor{gray!30}  % Set the background color of the second row
    InstructBLIP~\cite{dai2023instructblip} & Flan-T5-xl & 75.7 & 135.9 & 36.2 & 38.1 & 98.2 & 28.9 & 105.8 & 67.7 \\
    + \AlgoName  &            & 76.0 & \underline{136.9} & 37.4 & 39.4 & 99.9 & 28.8 & 106.5 & 69.2 \\
    $\Delta$     &            & {\color{OliveGreen}\textbf{+0.3}} & {\color{OliveGreen}\textbf{+1.0}} & {\color{OliveGreen}\textbf{+1.2}} & {\color{OliveGreen}\textbf{+1.3}} & {\color{OliveGreen}\textbf{+1.7}} & {\color{BrickRed}\textbf{-0.1}} &
    {\color{OliveGreen}\textbf{+0.7}} & {\color{OliveGreen}\textbf{+1.5}}\\
    \hline
    \rowcolor{gray!30}  % Set the background color of the second row
    InstructBLIP~\cite{dai2023instructblip} & Flan-T5-xxl & 76.1 & 136.1 & 37.4 & 38.7 & 99.0  & 31.1 & 106.1 & 68.5 \\
    + \AlgoName  &             & 76.5 & \textbf{138.2} & 38.4 & 40.0 & 101.7 & 30.7 & 107.4 & 70.5 \\
    $\Delta$     &             & {\color{OliveGreen}\textbf{+0.4}} & {\color{OliveGreen}\textbf{+2.1}} & {\color{OliveGreen}\textbf{+1.0}} & {\color{OliveGreen}\textbf{+1.3}} & {\color{OliveGreen}\textbf{+2.7}} & {\color{BrickRed}\textbf{-0.4}} &
    {\color{OliveGreen}\textbf{+1.3}} & {\color{OliveGreen}\textbf{+2.0}}\\
    \hline
    \rowcolor{gray!30}  % Set the background color of the second row
    LLaVA-1.5~\cite{liu2023improved}     & Vicuna-7B & \underline{79.7} & 133.5 & \underline{57.4} & \underline{61.6} & 126.4 & 33.9 & 106.6 & \underline{93.0} \\
    + \AlgoName   &           & \textbf{80.5} & 134.7 & \textbf{59.1} & \textbf{62.4} & \textbf{128.7} & \textbf{36.5} & \textbf{107.6} & \textbf{94.7}\\
    $\Delta$      &           & {\color{OliveGreen}\textbf{+0.8}} & {\color{OliveGreen}\textbf{+1.2}} & {\color{OliveGreen}\textbf{+1.7}} & {\color{OliveGreen}\textbf{+0.8}} & {\color{OliveGreen}\textbf{+2.3}} & {\color{OliveGreen}\textbf{+2.6}} &
    {\color{OliveGreen}\textbf{+1.0}} & {\color{OliveGreen}\textbf{+1.7}}\\
    \bottomrule
  \end{tabular}
  }
  \egroup
  \caption{\textbf{\AlgoName results}. 
  % v1
  % Evaluation of \AlgoName implemented into ViT+T5, BLIP2, InstructBLIP, and LLaVA-1.5 using multiple model sizes, trained on the data described in \cref{sec:data}. The analysis is focused on general VL benchmarks, scene-text ones, and 0-shot capabilities. \AlgoName consistently outperforms the different baselines by a substantial margin.
  % v2
  Quantitative comparison of \AlgoName integrated into ViT+T5, BLIP2, InstructBLIP, and LLaVA-1.5, using different model sizes, with these baselines trained on the data described in \cref{sec:data}. The evaluation covers general and scene-text VL benchmarks and 0-shot capabilities. \AlgoName consistently outperforms the different baselines, demonstrating its effectiveness and versatility.
  }
  \vspace{-2mm}
  \label{tab:qavit}
\end{table*}

\subsection{\AlgoName Performance Gains}
We evaluate \AlgoName on general (VQA$^\text{v2}$ and COCO) and scene-text (VQA$^\text{T}$, VQA$^\text{ST}$ and TextCaps) benchmarks, in addition to zero-shot setting (VizWiz~\cite{bigham2010vizwiz}).
Additionally, we calculate average scores by assigning equal weight to both visual question answering and image captioning tasks.

\noindent\paragraph{ViT+T5}
\vspace{-4mm}
First, we examine a simple yet effective approach -- a frozen CLIP\footnote{\url{https://huggingface.co/openai/clip-vit-large-patch14-336}}~\cite{radford2021learning} and Flan-T5~\cite{flan-t5} of different sizes (\texttt{base}, \texttt{large}, and \texttt{xl}), with an MLP projection module.
We train the system on the data described in \cref{sec:data}, using both the standard CLIP-ViT and \AlgoNameNoSpace, with the same training hyperparameters. 
In particular, we adapt the LLM weights using LoRa~\cite{hu2021lora}, train the projection MLP, and, in the \AlgoName case, also the instruction fusing counterparts.
Both the baseline and the \AlgoName settings exhibit high parameter efficiency, keeping the vast majority of the weights frozen.
We report the quantitative results of the ViT+T5 and compare them with \AlgoName in \Cref{tab:qavit}.
As can be seen, \AlgoName leads to a substantial and consistent improvement compared to the baseline in all the benchmarks and across all model sizes. 
Moreover, our method not only improves performance on the seen benchmarks, but it also benefits it in a zero-shot setting on VizWiz~\cite{bigham2010vizwiz}.

To better understand the gains achieved by \AlgoNameNoSpace, we provide qualitative results in the ViT+T5-large model in \cref{fig:exmaples}.
As seen, \AlgoName leads to better performance, specifically on image-question pairs that require reasoning over nuanced low-level details inside the image. For example, the image-question pair on the right requires focusing on the board, which is relatively small and marginal in importance compared to the entire image. Similar behavior is observed throughout all such examples.

\noindent\paragraph{State-of-the-art Models}
\vspace{-4mm}
After validating the efficacy of \AlgoName in a pretraining-free setting, we turn to experiment with already-trained leading VL models.
In this setting, we finetune the base model with and without \AlgoName using our training data introduced in \cref{sec:data}.
As in the ViT+T5 case, we employ a similar training setting by applying LoRa to the LLM and tuning the projection model and the \AlgoName components, if applicable.
Specifically, we consider BLIP2~\cite{li2023blip}, InstructBLIP~\cite{dai2023instructblip}, using different sizes, and LLaVA-1.5~\cite{liu2023improved}, top-performing multimodal architectures, and report the results in \cref{tab:qavit}.
As can be seen, \AlgoName consistently improves the baselines in all the tested architectures and across all the seen benchmarks while showing benefit also in the unseen one (except in InstructBLIP).

\begin{table}[t]
  \centering
  \bgroup
  \def\arraystretch{1.1}
      \resizebox{1\linewidth}{!}{%
      \begin{tabular}{lllll}
        \toprule
        \textbf{Method} & $\text{VQA}^{\text{v2}}$ & $\text{VQA}^{\text{T}}$ & TextCaps & VizWiz\\
        \hline
        mPLUG-DocOwl~\cite{ye2023mplug}    & -    & 52.6$^*$ & 111.9$^*$ & -   \\
        % BLIP2-xl        & 63.0 & - & - & 29.8 \\
        BLIP2~\cite{li2023blip}       & 65.0 & 23.4 & 70.4 & 29.4 \\
        % InstructBLIP-xl & -    & - & - & 32.7 \\
        InstructBLIP~\cite{dai2023instructblip}& -    & 30.9 & 75.6$^*$ & 30.9 \\
        InstructBLIP$^\text{+OCR}$~\cite{dai2023instructblip}& -    & 46.6 & 126.0$^*$ & 30.9 \\
        OpenFlamingo-9B~\cite{awadalla2023openflamingo} & 50.3 & 24.2 & - & 17.7  \\
        IDEFICS-9B~\cite{laurencon2023obelics}      & 50.9 & 25.9 & 25.4 & 35.5  \\    
        IDEFICS-80B~\cite{laurencon2023obelics}     & 60.0 & 30.9 & 56.8 & \underline{36.0}\\
        Shikra~\cite{chen2023shikra}          & 77.4$^*$ & - & - & -\\
        Qwen-VL~\cite{bai2023qwen}         & 79.5$^*$ & \textbf{63.8}$^*$ & - & 35.2 \\
        \hline
        \rowcolor{gray!30}  % Set the background color of the second row
        LLaVA-1.5~\cite{liu2023improved}       & \underline{79.7}$^*$ & 57.4$^*$ & \underline{126.4}$^*$ & 33.9 \\
        + \AlgoName     & \textbf{80.5}$^*$ & \underline{59.1}$^*$ &\textbf{128.7}$^*$ & \textbf{36.5} \\
        $\Delta$        & 
        {\color{OliveGreen}\textbf{+0.8}} & {\color{OliveGreen}\textbf{+1.7}} & {\color{OliveGreen}\textbf{+2.3}} &{\color{OliveGreen}\textbf{+2.6}} \\
        \bottomrule
      \end{tabular}
  }
  \egroup
  \caption{\textbf{Comparison to generalist models}. 
  Results comparison of \AlgoName integrated into LLaVA-1.5 with top-performing generalist models on VQA and captioning.
  \AlgoName outperforms existing methods in the VQA$^\text{v2}$, TextCaps and VizWiz. 
  Models marked with $^\text{+OCR}$ receive a list of OCR tokens, and scores noted with $^*$ signify that the dataset's training images are observed in training.
  }
  \vspace{-2mm}
  \label{tab:sota_comp}
\end{table}

\subsection{\AlgoName Results Analysis}
We turn to conduct a more in-depth analysis of the results provided in \cref{tab:qavit} to better understand the contributions of \AlgoNameNoSpace.
Our method improves the performance of different architectures, highlighting the three-way model agnosticism of \AlgoName in terms of the vision encoder, projection module, and LLM.
\begin{itemize}[nolistsep,leftmargin=*]
    \item \textbf{Vision Encoder} -- Despite BLIP2 and InstructBLIP utilizes a different vision encoder than LLaVA-1.5 ($39$-layered EVA-CLIP~\cite{fang2023eva} with a resolution of $224\times224$ vs. a $24$-layered CLIP ViT-L of $336\times336$ resolution), integrating \AlgoName leads to improved performance.
    
    \item \textbf{Projection Module} -- On the one hand, BLIP2 and InstructBLIP use a QFormer, a transformer-based architecture with learnable tokens, that also reduces the sequence length of the visual features by processing the different visual features. On the other hand, LLaVA-1.5 and ViT+T5 utilize a simple MLP that operates separately on the visual features. Despite this crucial difference, our method is compatible with both, leading to consistent gains.
    
    \item \textbf{LLM Architecture} -- We experiment with both encoder-decoder (FLAN-T5~\cite{flan-t5}) and decoder-only (Vicuna~\cite{vicuna2023}). In the encoder-decoder case, we encode the textual guidance using the preexisting encoder, and in the decoder-only, we utilize the model's embedding module.
    We provide a comparison between these two alternatives in \cref{sec:design_ablation}.
    Our experiments show that despite the significant LLM architecture differences, \AlgoName is compatible with both, showcasing its versatility.
\end{itemize}

Next, we examine the effects of scale-up on our approach by comparing the results of different model sizes.
In particular, we consider \texttt{base}, \texttt{large}, and \texttt{xl} and \texttt{xl} and \texttt{xxl} for ViT+T5 and BLIP2 and InstrucrtBLIP, respectively.
Our quantitative analysis demonstrates that our approach leads to consistent improvement across all model scales, making it compatible with different LLM sizes.
Remarkably, for a given LLM size, applying \AlgoName is more beneficial than scale-up in terms of average general and scene-text performance.
For example, InstructBLIP-xl + \AlgoName leads to $106.5$ and $69.2$ (general and scene-text averages), compared to InstructBLIP-xxl with $106.1$ and $68.5$ -- an improvement of \textcolor{OliveGreen}{$\mathbf{+0.4}$} and \textcolor{OliveGreen}{$\mathbf{+0.7}$}, compared to the scale-up.
Based on these results, we conduct a more thorough analysis of our method's contribution in \cref{sec:ocr}.

Lastly, we focus on InstructBLIP, as it utilizes an instruction-aware QFormer. In particular, this component processes the visual features with respect to the provided text, which conceptually resembles \AlgoNameNoSpace.
Thus, one might presume that utilizing such a model might make \AlgoName contribution redundant.
However, it is fundamentally different as our method is integrated inside the ViT and not on top of it.
Hence, the QFormer cannot compensate for information disregarded in the output features of the ViT.
On the contrary, \AlgoNameNoSpace, by being integrated into the ViT layers, can emphasize the relevant features and prevent their potential disregardance, leading to performance gains.

\subsection{Comparison to State-of-the-art}
Despite \AlgoName being a model-agnostic approach that can be integrated into any VL model, we compare LLaVA-1.5 + \AlgoName to other state-of-the-art generalist methods.
In particular, we consider mPLUG-DocOWL~\cite{ye2023mplug}, OpenFlamingo-9B~\cite{awadalla2023openflamingo}, IDEFICS-9B and 80B~\cite{laurencon2023obelics}, Shikra~\cite{chen2023shikra} and Qwen-VL~\cite{bai2023qwen}, and report the results in \cref{tab:sota_comp}.
As can be seen, \AlgoName pushes the performance of the LLaVA-1.5 model on the unseen VizWiZ beyond Qwen-VL and IDEFICS-80B, 
leading to the best performance across the considered models.
In addition, \AlgoName leads to the top-performing generalist model in VQA$^\text{v2}$.
%, outperforming Qwen-VL, which utilizes a higher resolution ($448 \times 448$) finetuned vision encoder.

\subsection{Why and When \AlgoName is Effective?}
\label{sec:ocr}

\begin{figure}[t]
    \centering
    \includegraphics[width=0.96\linewidth]{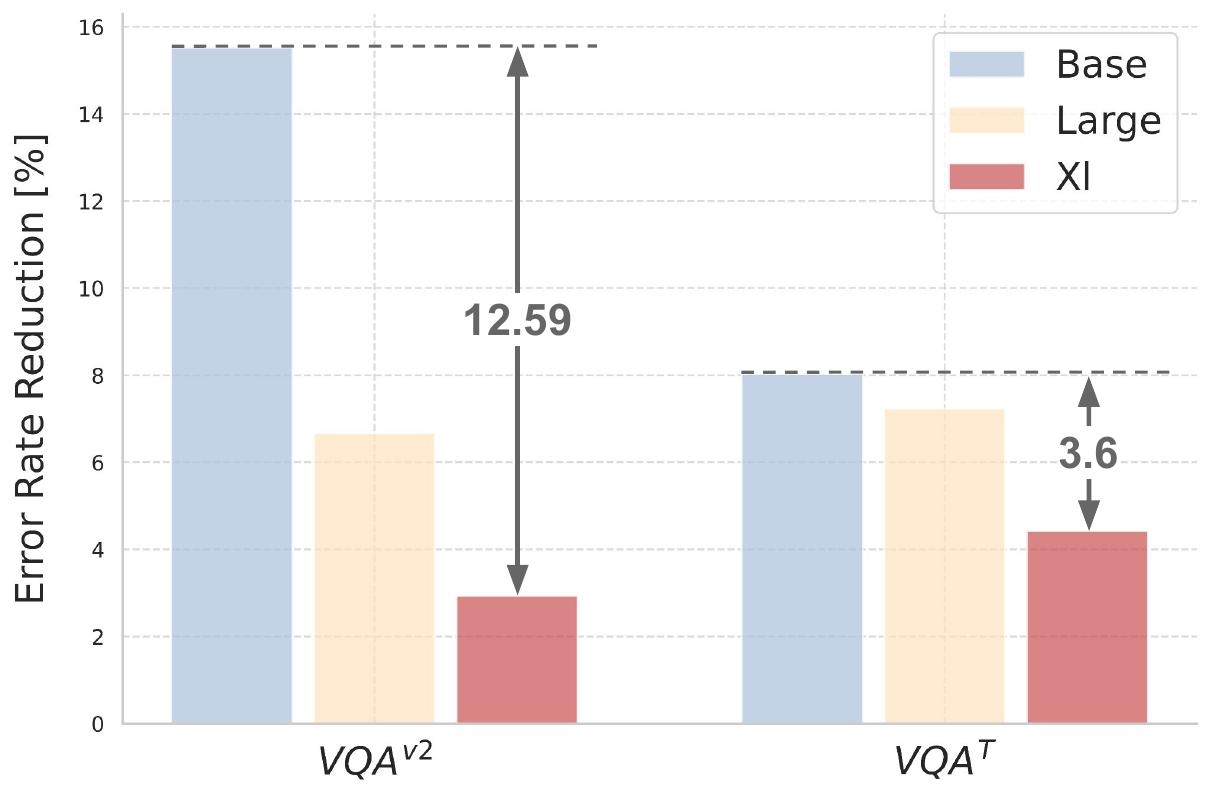}
    \caption{\textbf{\AlgoName effectiveness analysis.}
   Comparison of the trends in error rate reduction of \AlgoName in VQA$^\text{T}$ and VQA$^\text{v2}$ as the language model is scaled up.
   The relative performance improvements of our approach are more consistent across model scales in the former.
   These trends are attributed to each dataset's different question types' composition, where VQA$^\text{T}$ exhibits more questions focusing on non-salient and overlooked elements.}
    \label{fig:barplot}
\end{figure}

In this section, we better study the impact of \AlgoNameNoSpace.
We argue that our method plays a crucial role in addressing two common image-question fail-cases within VL architectures: first, questions regarding image aspects disregarded by the vision model, and second, questions related to elements encoded by the vision model but misinterpreted by the LLM.
While scaling up the LLM might mitigate some of the latter type of fail-case, the former remains challenging to address, hence, we consider the first as a more interesting setting for our method.
% v1
To examine our claim, we propose to compare the gains of \AlgoName across different LLM scales in two datasets, VQA$^\text{T}$ and VQA$^\text{v2}$, that differ in the composition of the fail-cases mentioned above.
We categorize VQA$^\text{T}$ as having more instances of the first fail-case and VQA$^\text{v2}$ as having more of the second one since OCR information is more likely to be disregarded due to its relative scarcity in the ViT's pretraining captions compared to non-OCR visual data.
Indeed, as anticipated, the trends in \cref{fig:barplot} align with our expectation that the gains of \AlgoName in VQA$^\text{T}$ would be more significant when scaling up compared to VQA$^\text{v2}$.
Although more substantial gains are generally observed in smaller models, our method leads to consistent improvements even on the largest models (\textit{i.e.}, BLIP2-xxl InstructBLIP-xxl and LLaVA-1.5), as evidenced in \cref{tab:qavit}.

\section{Ablation Studies}
\vspace{-1mm}
\label{sec:ablation}
In this section, we conduct extensive experiments to understand the performance improvements better and analyze the impact of our method.
We first study the effect of different design choices (\cref{sec:design_ablation}) and then analyze the contributions of different training data compositions (\cref{sec:ablation_data}).
Throughout this section, we focus on ViT-T5-large architecture.

\subsection{Design Choices}
\label{sec:design_ablation}
\vspace{-1mm}
We analyze different design choices and explore different settings for the textual guidance encoding and representations fusing while applying \AlgoNameNoSpace.
\begin{table}[t]
  \centering
  \bgroup
  \def\arraystretch{1.1}
  \resizebox{1\linewidth}{!}{%
  \begin{tabular}{llc|ll}
    \toprule
    Inst. & Fuse & Freeze & $\text{VQA}^{\text{v2}}$ & $\text{VQA}^{\text{T}}$ \\
    \hline
    \rowcolor{gray!30}
    \xmark & \xmark & \cmark      & 70.0 & 44.7 \\
    P.T. & \texttt{late} & \cmark & 70.1 (\textcolor{LimeGreen}{+0.1\%}) & 45.8 (\textcolor{LimeGreen}{+1.1\%})\\
    \xmark & \xmark & \xmark & 69.5 (\textcolor{BrickRed}{-0.5\%}) & 44.9 (\textcolor{LimeGreen}{+0.2\%})\\
    \cdashline{1-5}
    Enc. & \texttt{early} & \cmark & 67.9 (\textcolor{BrickRed}{-2.1\%}) & 41.7 (\textcolor{BrickRed}{-3.0\%})\\
    Enc. & \texttt{sparse} & \cmark & 70.7 (\textcolor{LimeGreen}{+0.7\%}) & 46.6 (\textcolor{LimeGreen}{+1.9\%})\\
    Enc. & \texttt{all} & \cmark & 69.5 (\textcolor{BrickRed}{-0.5\%}) & 45.9 (\textcolor{LimeGreen}{+1.2\%})\\
    \cdashline{1-5}
    Emb. & \texttt{late} & \cmark & 71.0 (\textcolor{LimeGreen}{+1.0\%}) & 47.5 (\textcolor{LimeGreen}{+2.8\%})\\
    BERT & \texttt{late} & \cmark & 71.8 (\textcolor{LimeGreen}{+1.8\%}) & 48.3 (\textcolor{LimeGreen}{+3.6\%})\\
    CLIP & \texttt{late} & \cmark & 71.8 (\textcolor{LimeGreen}{+1.8\%}) & 48.0 (\textcolor{LimeGreen}{+3.3\%})\\
    \hline
    \rowcolor{golden}  % Set the background color of the second row
    Enc. & \texttt{late} & \cmark & \textbf{72.0} (\textcolor{OliveGreen}{\textbf{+2.0\%}}) & \textbf{48.7} (\textcolor{OliveGreen}{\textbf{+4.0\%}})\\
    % \cdashline{1-6}
    % Pre. & \texttt{late} & IN & \cmark & - & -\\
    \bottomrule
  \end{tabular}
  }
  \egroup
  \caption{
  % v1
  % \textbf{\AlgoNameNoSpace 's design choices}. Comparison of the effects of \AlgoName different configurations on the VQA$^\text{v2}$ and VQA$^\text{T}$ results.
  % v2
  \textbf{Design choices ablation}.
  We mark the baseline and our top-performing configuration of \AlgoName in grey and yellow, respectively.
  Top: Results of different finetuning strategies.
  Middle: The effect of different integration points of \AlgoNameNoSpace.
  Bottom: Comparison of different instruction (Inst.) encodings.
  }
  \label{tab:comp_ablation}
  \vspace{-4mm}
\end{table}
\vspace{-4mm}
\paragraph{Finetuning Strategy}
Despite being parameter efficient, \AlgoName introduces more trainable parameters than the baseline.
To validate that the improvements are credited to the method and not the additional capacity, we conduct experiments with two other finetuning techniques.
First, analogous to deep prompt tuning, we train our model while inserting into \AlgoName a fixed textual prompt instead of the relevant question.
By employing the same blocks as our method, this interpretation of prompt tuning (denoted as P.T.) isolates the contribution of question-conditioned image encoding. In addition, we also experiment with finetuning the entire baseline's vision encoder, which introduces a significant amount of trainable parameters.
The results in the top part of \cref{tab:comp_ablation} show that while \AlgoName leads to ${+2.0\%}$ and ${+4.0\%}$ on VQA$^\text{v2}$ and VQA$^\text{T}$, P.T improves solely in ${+0.1\%}$ and ${+1.1\%}$, respectively.
Comparing \AlgoName results with P.T. enables decomposing our method's improvement into gains attributed to additional capacity and to question-aware visual features, implying that the latter is the most significant.
In addition, full finetuning CLIP, which introduces training instability, improves the baseline in VQA$^\text{T}$ but reduces it on VQA$^\text{v2}$.
This supports the choice of current VL works to freeze the ViT during pretraining.

\vspace{-4mm}
\paragraph{Integration Point}
We explore different fusing locations -- \texttt{early} (bottom layers), \texttt{late} (top layers), \texttt{sparse} (every $2$ layers), and \texttt{all} (every layer).
While \texttt{early}, \texttt{sparse}, and \texttt{late} add the same amount of trainable parameters, \texttt{all} doubles it.
The results presented in the middle part of \cref{tab:comp_ablation} demonstrate the significant advantage of \texttt{late} fusion.
We attribute this to the hierarchical structure of the ViT's layers, in which early layers specialize in capturing low-level and localized visual details, while higher ones focus on extracting more abstract and high-level visual features.
Thus, disregarding question-related image aspects is more likely to occur on the higher layers, \AlgoName is most effective in \texttt{late} fusion.
Moreover, as the early layers extract low-level details, they should not be modified, and applying \AlgoName to them impairs the results.

\paragraph{Question Representation}
\vspace{-4mm}
As specified in \cref{sec:method}, we use the preexisting LLM's encoder (Enc.) to obtain the question representation.
Here, we study the effect of different such choices and present their results at the bottom of \cref{tab:comp_ablation}.
First, utilizing solely the embeddings (Emb.) is less effective than the encoder.
We attribute this to the improved contextual understanding of the latter, enabling better guidance to the visual features in \AlgoName. 
Next, we experiment with using a designated language model, considering both a BERT~\cite{devlin2018bert} and the corresponding CLIP text encoder.
While utilizing the system's language model is more parameter efficient and can lead to more seamless integration, a dedicated language model can better align with the vision model and offer a more modular and generic design.
As can be seen, while both perform satisfactorily, the designated LLM is superior, while BERT outperforms CLIP.

\subsection{The Impact of Training Data}
\vspace{-1mm}
\label{sec:ablation_data}
\begin{table}[t]
  \centering
  \bgroup
  \def\arraystretch{1.1}
      \resizebox{1\linewidth}{!}{%
  \begin{tabular}{lc|cccc}
    \toprule
    Datasets & Size & $\text{VQA}^{\text{v2}}$ & $\text{VQA}^{\text{T}}$ & COCO & TextCaps \\
    \hline
    VQA    & 2.3M & 71.2  & 45.8  & 29.9 & 34.3 \\
    + CAP  & 3.0M & 71.5  & 47.4  & 117.5   & 106.1 \\
    + DOC  & 3.1M & \textbf{72.0} & \textbf{48.7} & \textbf{118.7} & \textbf{106.2} \\
    \bottomrule
  \end{tabular}
  }
  \egroup
  \caption{\textbf{Training data ablation}. Contribution analysis of different training dataset compositions on visual question answering and captioning, demonstrating the importance of multi-task data.}
  \vspace{-3.5mm}
  \label{tab:data_ablation}
\end{table}
Our training data, described in \cref{sec:data}, consists of three main data types: i) natural images visual question answering (VQA); ii) natural image captioning (CAP); and iii) documents understanding (DOC).
We turn to evaluate the contribution of each of them and report the results in \cref{tab:data_ablation}.
As can be seen, adding CAP datasets into the VQA ones (second row) not only improves the captioning performance but also boosts the performance on the VQA ones. We attribute this to the enlargement and diversification of the training data. Moreover, incorporating DOC data, despite the significant change of domain (natural images vs. documents), increases the performance. 
We hypothesize that this is because \AlgoName maintains the original visual capabilities; it prevents the performance drop due to multi-domain data while leading to better OCR understanding. 
This, in return, improves the overall results, as observed in ~\cite{ganz2023towards}.

\section{Discussion and Conclusions}
\vspace{-1mm}
In this work, we introduced an approach to condition the vision encoder in any multimodal vision-language architecture, named \AlgoNameNoSpace. Our method leads to question-aware visual features, improving their alignment with the provided query. Through extensive experimentation across a diverse set of vision-language models, we have demonstrated the effectiveness and versatility of our method. It consistently enhances the performance of these models across a range of benchmark tasks, encompassing both general and scene-text domains, as well as the challenging zero-shot setting. The introduction of \AlgoName represents a notable advancement in the pursuit of question-aware vision within VL modeling, making models more context-aware and enabling them to excel in various tasks. 
We hope our method will inspire further research striving towards improved text-aware mechanisms and designated pretraining techniques.

\clearpage

{
    \small
    \bibliographystyle{ieeenat_fullname}
    \bibliography{main}
}

\clearpage
\setcounter{page}{1}
\maketitlesupplementary

\appendix

\section{Implementation Details}
\paragraph{Overall Training Protocol}
% \vspace{-5pt}
% \textbf{Overall Training Protocol}
For all of the considered architectures, we follow the same general training procedure in which we apply LoRa~\cite{hu2021lora} to the LLM and finetune the projection module. When applying \AlgoNameNoSpace, we also finetune the instruction representation projection MLPs.
In particular, we employ LoRa ($\alpha$=32, r=16, dropout=0.05, and the queries and keys as the target modules) and utilize an AdamW~\cite{loshchilov2018fixing} optimizer ($\beta_1, \beta_2 = 0.9, 0.999$ and ${\epsilon = 1e-08}$) with cosine annealing scheduler~\cite{loshchilov2016sgdr} that decays to $\times 0.01$ from the base learning rate.
In addition, we perform $1000$ warm-up steps.
We use 8 Nvidia A100 (40G) GPUs in all of our experiments with bfloat16.
Next, we provide the specific implementation details regarding ViT+T5, BLIP2, InstructBLIP, and LLaVA-1.5.

% \vspace{-15pt}
\paragraph{ViT+T5}
% \noindent\textbf{ViT+T5}
ViT+T5 is comprised of a CLIP~\cite{radford2021learning} ViT-L vision encoder that operates in a $336\times336$ resolution, coupled with a FLAN-T5 encoder-decoder model~\cite{flan-t5} using an MLP projection module.
The projection component consists of two linear layers that map from the ViT's dimension $D_1$ into the LLM's one $D_2$ ($D_1 \rightarrow D_2 \rightarrow D_2$).
We train three variants of ViT+T5, which differ in the LLM scale, where we consider \texttt{base}, \texttt{large}, and \texttt{xl}.
We use the LLM's encoder as the question encoder and train the models on our multi-task dataset (\cref{sec:data}) for $5$, $2$, and $2$ epochs, using a batch size per GPU of $16$, $8$, and $6$, with a learning rate of $1e-4$, $5e-5$ and $1e-5$, respectively.
\AlgoName introduces 38M, 45M, and 66M trainable parameters out of the overall 589M, 1,132M, and 3,220M.
In addition, when applying \AlgoName to a pretraining-free setup, we observe that using a higher learning rate ($\times 100$) for the projection module stabilizes the training.
We hypothesize that while the vision encoder and LLM are pretrained separately, the projection module is randomly initialized, and thus, its weights should be adjusted more than the former counterparts.

% \vspace{-15pt}
\paragraph{BLIP2 and InstructBLIP}
% \noindent\textbf{BLIP2 and InstructBLIP}
We experiment with both the \texttt{xl} and \texttt{xxl} models, and similar to the ViT+T5, we use the LLM's encoder for processing the question before feeding it into \AlgoName. We use a single learning rate group for all models for all the trainable parameters.
For the \texttt{xl} models, we train for $2$ epochs, with a batch size of $8$ per GPU with a base learning rate of $2e-5$.
For the \texttt{xxl} ones, we reduce the batch size to $4$ per GPU. 
In addition, we employ a weight decay of $0.05$ for all models.

% \vspace{-15pt}
\paragraph{LLaVA-1.5}
% \noindent\textbf{LLaVA-1.5}
As LLaVA-1.5 is based on a decoder-only LLM, we use the model's embedding module to process the questions when applying \AlgoName. We train for one epoch with an effective batch size of $4$ per GPU (using 2-step gradient accumulation) and a base learning rate of $5e-5$.

\begin{table}[t]
  \centering
  \bgroup
  \def\arraystretch{1.1}
      \resizebox{1\linewidth}{!}{%
        \begin{tabular}{l}
        \toprule
        Template \\
        \hline
        $<$image$>$"A short image caption:" \\
        $<$image$>$"A short image description:" \\
        $<$image$>$"A photo of" \\
        $<$image$>$"An image that shows" \\
        $<$image$>$"Write a short description for the image." \\
        $<$image$>$"Write a description for the photo." \\
        $<$image$>$"Provide a description of what is presented in the photo." \\
        $<$image$>$"Briefly describe the content of the image." \\
        $<$image$>$"Can you briefly explain what you see in the image?" \\
        $<$image$>$"Could you use a few words to describe what you perceive in the photo?" \\
        $<$image$>$"Please provide a short depiction of the picture."\\
        $<$image$>$"Using language, provide a short account of the image."\\
        $<$image$>$"Use a few words to illustrate what is happening in the picture."\\
        \bottomrule
      \end{tabular}
  }
  \egroup
  \caption{\textbf{Captioning instruction templates}. The instruction templates used for the captioning datasets. For VQA, we simply use the provided question.
  }
  \label{tab:cap_inst}
  % \vspace{-15pt}
\end{table}

% \vspace{-5pt}
\section{Multi-Task Training Dataset and Evaluation}
% \vspace{-5pt}
As stated in \cref{sec:data}, we utilize a multi-task dataset that contains multiple benchmarks of different tasks. In \cref{tab:data}, we provide a detailed list of the training datasets and the evaluation metric and split used for reporting results throughout the paper.

\begin{table*}[t]
% \vspace{-10pt}
  \centering
  \bgroup
  \def\arraystretch{1.1}
      \resizebox{1\linewidth}{!}{%
      % \begin{tabular}{l|l|l|ll|l}
      %   \toprule
        % \multirow{2}{*}{Task} & \multirow{2}{*}{Dataset} & \multirow{2}{*}{Description} & \multicolumn{2}{c|}{Split} & \multirow{2}{*}{Metric} \\
        \begin{tabular}{l|l|l|l|l}
        \toprule
        Task & Dataset & Description & Eval split & Metric \\
        \hline
        Image Caption & COCO & Captioning of natural images & karpathy-test & CIDEr($\uparrow$) \\
        \hline
        Scene-Text Caption & TextCaps & Text-oriented captioning of natural images & validation & CIDEr($\uparrow$) \\
        \hline
        \multirow{2}{*}{General VQA} & VQA$^\text{v2}$ & VQA on natural images & test-dev & vqa-score($\uparrow$)\\
        & Visual Genome & VQA on natural images & - & - \\
        \hline
        \multirow{3}{*}{Scene-Text VQA} & VQA$^\text{T}$ & Text-oriented VQA on natural images & validation & vqa-score($\uparrow$) \\
        & VQA$^\text{ST}$ & Text-oriented VQA on natural images & test & ANLS($\uparrow$)\\
        & VQA$^\text{OCR}$ & Text-oriented VQA on book covers & - & - \\
        \hline
        \multirow{3}{*}{Documents Understanding} & DocVQA & VQA on scanned documents & test & ANLS($\uparrow$) \\
        & InfoVQA & VQA on infographic images & test & ANLS($\uparrow$)\\
        & ChartQA & VQA on chart images & -&-\\
        \bottomrule
      \end{tabular}
  }
  \egroup
  \caption{\textbf{Training datasets and evaluation}. The datasets used for training alongside their evaluation split and metric, if applicable.
  }
  \label{tab:data}
  % \vspace{-10pt}
\end{table*}

% \vspace{-5pt}
\section{Image Captioning Templates}
% \vspace{-5pt}
For the VQA-based datasets, we simply utilize the provided question to guide \AlgoNameNoSpace. 
However, in the captioning case, it is infeasible.
Thus, we use the captioning templates used in InstructBLIP~\cite{dai2023instructblip} and provide them in \cref{tab:cap_inst} for completeness.
These captions are sampled uniformly during training and inference.

\begin{table*}[t]
  \centering
  % \bgroup
  % \def\arraystretch{1.}
  \resizebox{1\linewidth}{!}{%
  \tiny
  \begin{tabular}{ll|ccc|ccc}
    \hline
    \multirow{2}{*}{\textbf{Method}} & \multirow{2}{*}{LLM} & \multicolumn{3}{c|}{\textbf{Scene-Text}} & \multicolumn{3}{c}{\textbf{Documents}}\\
    & & $\text{VQA}^{\text{T}}$ & $\text{VQA}^{\text{T}}_{\text{Read}}$ &  $\text{VQA}^{\text{T}}_{\text{See}\boldsymbol{\cap}\text{Read}}$ & DocVQA & InfoVQA & Average \\
    % \hline
    % \rowcolor{gray!30}  % Set the background color of the second row
    % ViT+T5-base & Flan-T5-base & 32.0 & 19.7 & 25.9 & 40.2 & 41.3 & 28.8\\
    % + \AlgoName & & 36.1 & 20.5 & 28.3 & 45.0 & 46.1 & 32.3\\
    % $\Delta$ & &  {\color{OliveGreen}\textbf{+4.1}} & {\color{OliveGreen}\textbf{+0.8}} & {\color{OliveGreen}\textbf{+2.4}}  & {\color{OliveGreen}\textbf{+4.8}} & {\color{OliveGreen}\textbf{+4.8}} & {\color{OliveGreen}\textbf{+3.5}}\\
    % \hline
    % \rowcolor{gray!30}  % Set the background color of the second row
    % ViT+T5-large  &Flan-T5-large & 36.7 & 22.1 & 29.4  & 44.7 & 46.0 & 32.0\\
    % + \AlgoName   && 39.4 & 21.8 & 30.6  & 48.7 & 50.0 & 36.7 \\
    % $\Delta$ && {\color{OliveGreen}\textbf{+2.7}} & {\color{BrickRed}\textbf{-0.3}} & {\color{OliveGreen}\textbf{+1.2}} & {\color{OliveGreen}\textbf{+4.0}} & {\color{OliveGreen}\textbf{+4.0}} & {\color{OliveGreen}\textbf{+4.7}}\\
    \hline
    \rowcolor{gray!30}  % Set the background color of the second row
    ViT+T5-xl   &Flan-T5-xl & 48.0 & 49.3 & 35.6 & 42.3 & 26.4 & 34.4\\
    + \AlgoName & & 50.3 & 51.8 & 36.2 & 44.2 & 27.1 & 35.7\\
    $\Delta$ && {\color{OliveGreen}\textbf{+2.3}} & {\color{OliveGreen}\textbf{+2.5}} & {\color{OliveGreen}\textbf{+0.6}} & {\color{OliveGreen}\textbf{+1.9}} & {\color{OliveGreen}\textbf{+0.7}} & {\color{OliveGreen}\textbf{+1.3}} \\
    % \Xhline{1pt}
    \hline
    \rowcolor{gray!30}  % Set the background color of the second row
    BLIP2      & Flan-T5-xl   & 34.5 & 36.1 & 18.7 & 16.1 & 21.1 & 18.6\\
    + \AlgoName   &           & 36.6 & 38.3 & 20.4 & 17.1 & 21.2 & 19.2  \\
    $\Delta$ && {\color{OliveGreen}\textbf{+2.1}} & {\color{OliveGreen}\textbf{+2.2}} & {\color{OliveGreen}\textbf{+1.7}} & {\color{OliveGreen}\textbf{+1.0}} & {\color{OliveGreen}\textbf{+0.1}} & {\color{OliveGreen}\textbf{+0.6}} \\
    % \hline
    % \rowcolor{gray!30}  % Set the background color of the second row
    % BLIP2-xxl  & Flan-T5-xxl & 17.0 & 23.5 & 20.3  & 36.5 & 38.2 & 20.7 \\
    % + \AlgoName             && 16.9 & 23.8 & 20.4  & 37.5 & 39.3 & 20.7 \\
    % $\Delta$ && {\color{BrickRed}\textbf{-0.1}} & {\color{OliveGreen}\textbf{+0.3}} & {\color{OliveGreen}\textbf{+0.1}} & {\color{OliveGreen}\textbf{+1.0}} & {\color{OliveGreen}\textbf{+1.1}} & {\color{gray}\textbf{0.0}}\\
    \hline
    \rowcolor{gray!30}  % Set the background color of the second row
    InstructBLIP  & Flan-T5-xl   & 36.2 & 37.9 & 19.3 & 17.3 & 19.9 & 18.6\\
    + \AlgoName                 && 37.4 & 39.0 & 22.5 & 18.2 & 20.5 & 19.3\\
    $\Delta$ & & {\color{OliveGreen}\textbf{+1.2}} & {\color{OliveGreen}\textbf{+1.1}}& {\color{OliveGreen}\textbf{+3.2}} & {\color{OliveGreen}\textbf{+0.9}} & {\color{OliveGreen}\textbf{+0.6}} & {\color{OliveGreen}\textbf{+0.7}} \\
    % \hline
    % \rowcolor{gray!30}  % Set the background color of the second row
    % InstructBLIP-xxl  & Flan-T5-xxl & 18.3 & 23.5 & 20.9  & 37.4 & 38.9 & 22.5\\
    % + \AlgoName      &              & 17.7 & 24.0 & 20.9  & 38.4 & 40.1 & 21.2 \\
    % $\Delta$ && {\color{BrickRed}\textbf{-0.5}} & {\color{OliveGreen}\textbf{+0.5}} & {\color{gray}\textbf{0.0}} & {\color{OliveGreen}\textbf{+1.0}} & {\color{OliveGreen}\textbf{+1.2}} & {\color{BrickRed}\textbf{-1.3}}\\
    \hline
    \rowcolor{gray!30}  % Set the background color of the second row
    LLaVa-1.5    & Vicuna-7B & 57.4 & 59.0 & 42.5 & 44.1 & 32.1 & 38.1 \\
    + \AlgoName  && 59.1 & 60.7 & 43.5 & 45.4 & 32.1 & 38.8 \\
    $\Delta$     && {\color{OliveGreen}\textbf{+1.7}} & {\color{OliveGreen}\textbf{+1.7}} & {\color{OliveGreen}\textbf{+1.0}} & {\color{OliveGreen}\textbf{+1.3}} & {\color{gray}\textbf{0.0}} & {\color{OliveGreen}\textbf{+0.7}}\\
    \hline
  \end{tabular}
  }
  % \egroup
  \caption{\textbf{Additional OCR Results.} Results on documents understanding and comprehensive VQA$^\text{T}$ analysis.}
  \label{tab:docs}
  % \vspace{-20pt}
\end{table*}

% \vspace{-5pt}
\section{Additional OCR Results}

\subsection{In-Depth Scene-Text analysis}
As explained in \cref{sec:ocr}, we view the scene-text benchmarks as an interesting testing bed for our approach.
To understand the contribution of \AlgoName for scene-text understanding, we follow the analysis of \citet{ganz2023towards} and decompose the results of VQA$^\text{T}$ into two non-overlapping subsets -- i) $\text{VQA}^{\text{T}}_{\text{See}\boldsymbol{\cap}\text{Read}}$ is the manually curated subset which contains questions that require reasoning over OCR and visual information simultaneously. We view this subset as the most challenging one. ii) VQA$^\text{T}_\text{Read}$ is composed of questions that can be answered solely by using the OCR information. 
The unification of these subsets results in the entire VQA$^\text{T}$ validation set.
We provide the results on these subsets on the middle section of \cref{tab:docs}. 
As can be seen, \AlgoName improves the results on VQA$^\text{T}_\text{Read}$ in all the models.
This highlights the ability of our method to better harness some of the overlooked OCR information.
In addition, it leads to consistent improvements on the $\text{VQA}^{\text{T}}_{\text{See}\boldsymbol{\cap}\text{Read}}$, which requires cross-modal reasoning over the OCR and visual cues.

% \vspace{-5pt}
\subsection{Documents Understanding}
% In this section, we report the results of both \AlgoName and the different baselines on document understanding using DocVQA and InfoVQA, alongside the average score on the middle part of \cref{tab:docs}.
% While the former contains questions regarding dense-text scanned documents, the latter is focused on reasoning over infographics.
% This is a highly challenging setting as documents constitutes a significant domain shift and CLIP is known to be highly limited in dense-text scenarios.
% Thus, applying \AlgoName which focuses existing limited visual features is not expected to lead a substantial boost.
% Yet, while the obtained results are far from state-of-the-art, \AlgoName consistently improves the baseline results, reflecting the capability of our method to focus the visual features on the OCR information.
% \vspace{-5pt}
In this section, we present the performance results of both \AlgoName and the various baseline models in the context of document understanding, evaluated on DocVQA and InfoVQA, as detailed in the right section of \cref{tab:docs}. 
DocVQA encompasses questions related to dense-text scanned documents, while InfoVQA is designed for reasoning over infographics. 
Operating in these domains is highly challenging as it constitutes a substantial domain shift for the CLIP vision encoder (from natural images to documents and inforgraphichs).
Moreover, as CLIP is inherently limited in dense-text scenarios, the application of \AlgoNameNoSpace, which specifically targets existing visual features, is not anticipated to yield a significant performance boost in such settings.
Despite these challenges, our results, while far from state-of-the-art levels, consistently demonstrate improvements over baseline performance. This underscores the effectiveness of our method in directing visual attention towards OCR information within the given constraints.

% \vspace{-5pt}
\section{Additional Qualitative Results and Analysis}

\begin{figure}
    \centering
    \includegraphics[width=0.95\linewidth]{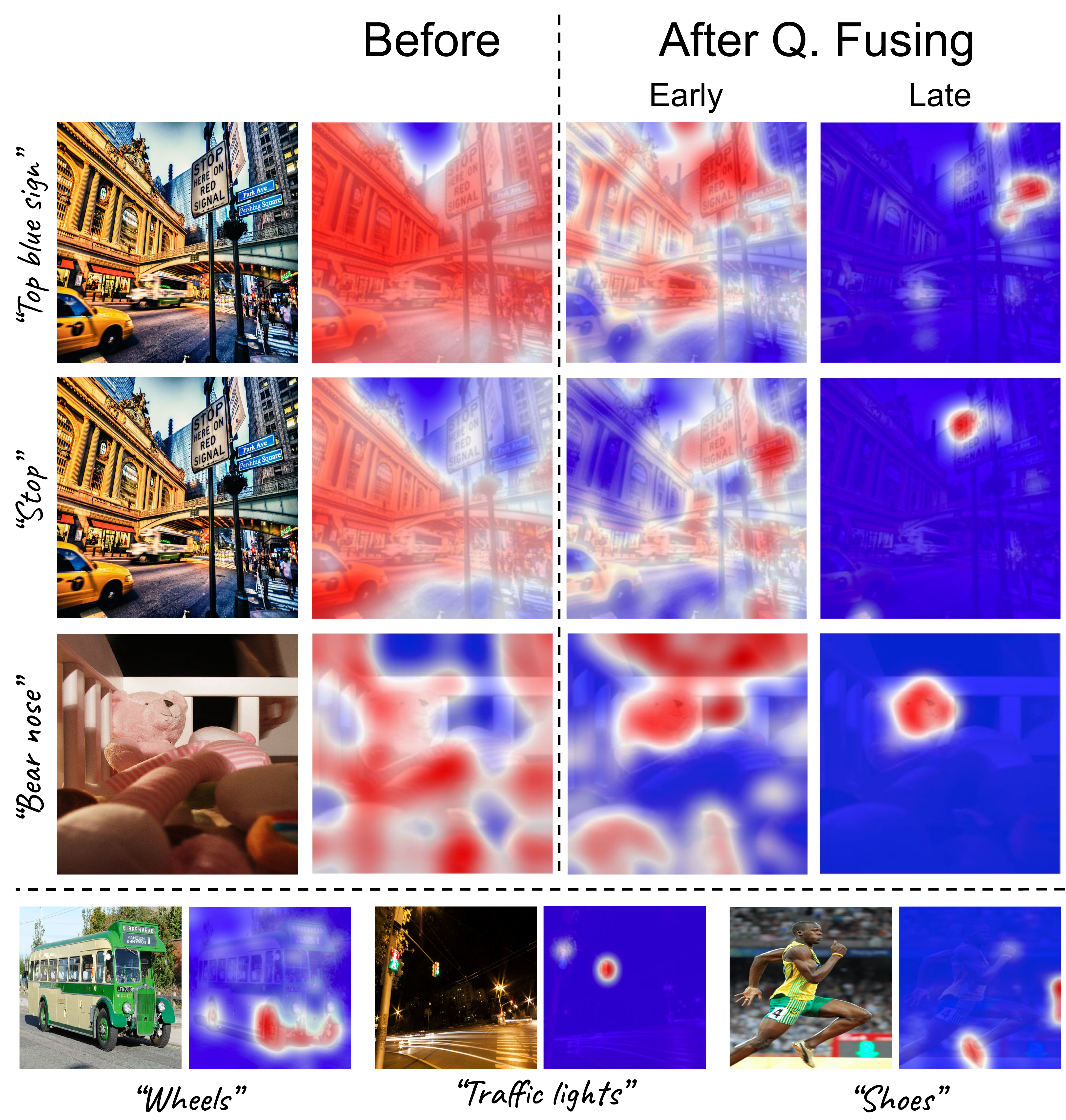}
    \caption{\textbf{Elaborated interpretations of QA-ViT.} 
    Additional visual and textual features interaction demonstrations, including visualizations at different granularity levels within the ViT.
    }
    \label{fig:layers_vis}
    % \vspace{-10pt}
\end{figure}

In \cref{fig:layers_vis}, we extend the visualizations conducted in the main paper to focus on the alignment of the text queries and visual features and provide additional demonstrations:
\begin{itemize}[leftmargin=*]   % nolistsep
    \item We provide attention visualizations at three levels of granularity within the ViT: (i) before the question fusing, (ii) immediately after it, and (iii) at the final layer.
    Illustrated in \cref{fig:layers_vis}, in (i), the network's attention spans across the entire visual content, while in (ii) and (iii), it focuses on fine-grained details according to the provided text. Specifically, the interaction of the text and vision throughout QA-ViT leads to more focused attention maps, as can be seen in the rightmost two columns.

    \item To better demonstrate the fine-grained interaction of text and vision in QA-ViT, we show the attention maps of the same image with respect to different text prompts (top two rows).
    This highlights QA-ViT's ability to shift the focus of the visual features based on the provided text.

    \item The bottom row contains additional visual-textual attention visualization, indicating QA-ViT's text-based focus. 
\end{itemize}

\begin{figure*}[t]
    \centering
    \includegraphics[width=\textwidth]{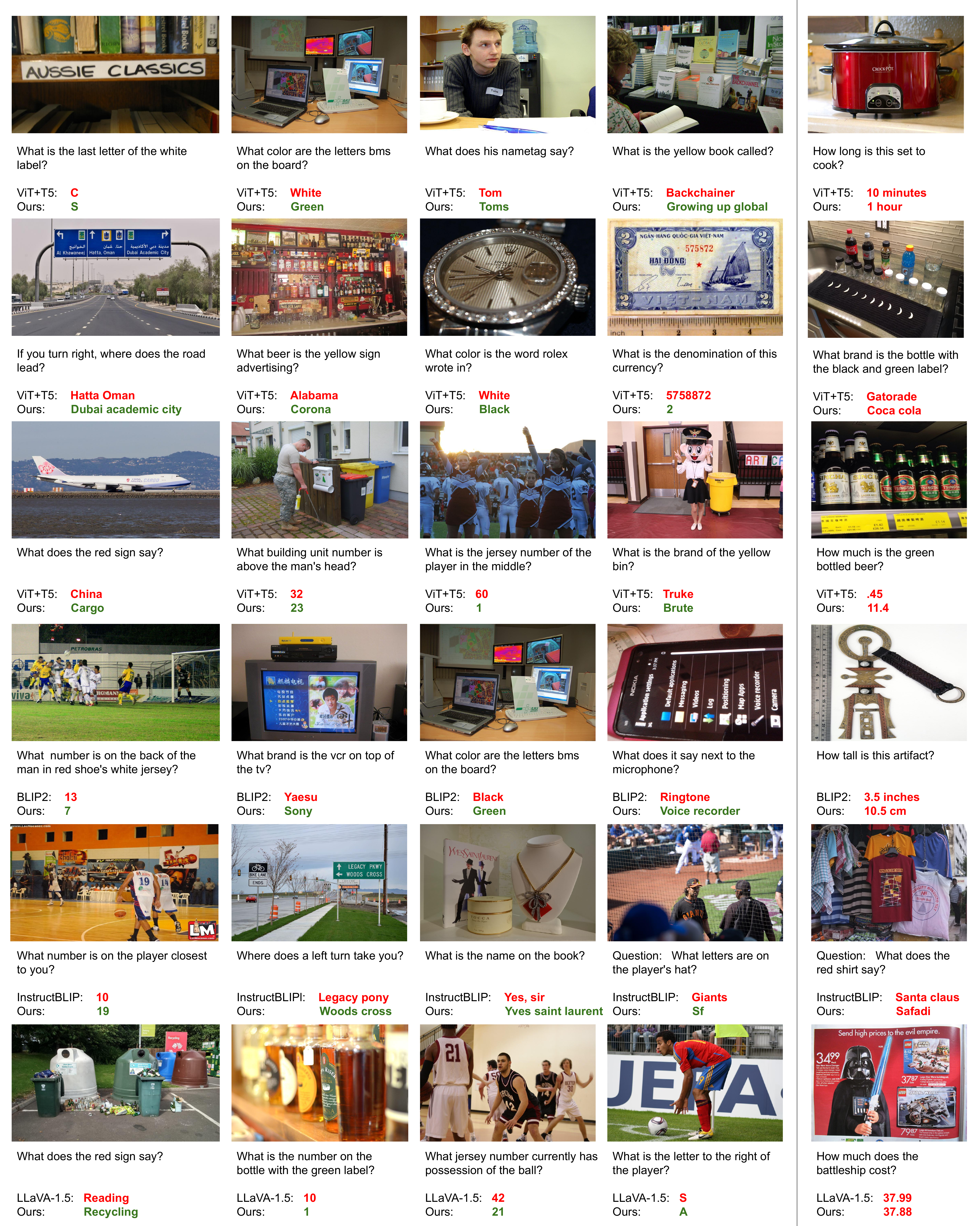}
    \caption{\textbf{Additional qualitative results.} Comparison between the baseline and our method on VQA$^\text{T}$ validation set using ViT+T5 (\texttt{base}, \texttt{large}, \texttt{xl}), BLIP2 and InstructBLIP (\texttt{xxl}) and LLaVA-1.5. Success and fail cases are presented on the left and right, respectively.
    }
    \label{fig:qulitative}
\end{figure*}

\noindent In addition, we provide qualitative comparison between QA-ViT and and the baseline in \cref{fig:qulitative}.  

% \begin{itemize}
%     \item results : see and read, docs
%     \item is it really question aware
%     \item data table
%     \item implementation details
%     \item table of added trainables
%     \item in depth zero shot vizwiz - our gain is not from unanswerable?
%     \item what are the used metrics
% \end{itemize}

\end{document}